\documentclass[runningheads]{llncs}
\usepackage[T1]{fontenc}
\usepackage{graphicx}
\usepackage{color}
\usepackage{hyperref}

\usepackage{multirow}
\usepackage{array}
\usepackage{diagbox}
\usepackage{makecell}
\usepackage{booktabs}
\usepackage{float}
\usepackage{fontawesome}
\usepackage{CJKutf8}
\usepackage{amsmath}
\usepackage{amssymb}

\begin{document}
	\title{Rotation-free Online Handwritten Character Recognition Using Linear Recurrent Units}
	\titlerunning{Rotation-free OLHCR using LRU}
	%
	
	\author{Zhe Ling, Sicheng Yu,\and Danyu Yang\textsuperscript{(\faEnvelopeO)}}
	
	\authorrunning{Zhe L. et al.}
	%
	\institute{College of Mathematics and Statistics, Chongqing University, China
	\email{202306021008@stu.cqu.edu.cn, 20201868@stu.cqu.edu.cn, danyuyang@cqu.edu.cn}\\ }
		
	\maketitle              

	\begin{abstract}
	Online handwritten character recognition leverages stroke order and dynamic features, which generally provide higher accuracy and robustness compared with offline recognition. However, in practical applications, rotational deformations can disrupt the spatial layout of strokes, substantially reducing recognition accuracy. Extracting rotation-invariant features therefore remains a challenging open problem. In this work, we employ the Sliding Window Path Signature (SW-PS) to capture local structural features of characters, and introduce the lightweight Linear Recurrent Unit (LRU) as the classifier. The LRU combines the fast incremental processing capability of recurrent neural networks (RNNs) with the efficient parallel training of State Space Models (SSMs), while reliably modeling dynamic stroke characteristics. We conducted recognition experiments with random rotation angle up to $\pm 180^{\circ }$ on three subsets of the CASIA-OLHWDB1.1 dataset: digits, English upper letters, and Chinese radicals. The accuracies achieved after ensemble learning were $99.62\%$, $96.67\%$, and $94.33\%$, respectively. Experimental results demonstrate that the proposed SW-PS+LRU framework consistently surpasses competing models in both convergence speed and test accuracy.

	\end{abstract}

\section{Introduction}
	With the widespread adoption of smart terminals, Online Handwritten Character Recognition (OLHCR) now plays a key role in applications ranging from office automation to education. By directly capturing temporal stroke trajectories, OLHCR exhibits stronger discriminative power than offline methods~\cite{liu2013online,yadav2015handwriting}. In many writing systems, characters are composed of semantic substructures—such as Chinese radicals—that play a central role in recognition. Accurately recognizing these components not only boosts overall recognition accuracy but also facilitates downstream tasks like structural decomposition and character retrieval~\cite{shi2003handwritten,ZHANG2020107305}. However, real-world factors like writing posture and device orientation frequently induce rotational deformations. Such deformations significantly alter stroke spatial layouts, causing confusion among structurally similar classes~\cite{5277580}. This challenge is particularly pronounced in rotation-sensitive models, which suffer from severe performance degradation due to the high structural similarity of rotated characters.
	
	While the path signature captures the geometry of a path~\cite{graham2013sparse}, its global application is hindered by its high dimensionality and insensitivity to local changes~\cite{yang2016rotation}. To address these limitations, we adopt the Sliding Window Path Signature (SW-PS) method~\cite{graham2013sparse}. By segmenting the trajectory into chronological windows and computing the fixed-order signature for each window, SW-PS effectively captures local dynamics while maintaining a manageable feature dimension. This approach balances global structure with local details, significantly enhancing robustness to rotational perturbations during training and inference.
	
	In recent years, State Space Models (SSMs)~\cite{gu2021efficiently} have performed exceptionally well in long sequence modeling, capable of handling sequences with linear computational complexity and effectively capturing global dependencies. Inspired by the design of state space models, the Linear Recurrent Unit (LRU) proposed by the DeepMind team~\cite{orvieto2023resurrecting} improves the stability and modeling capabilities of traditional RNNs~\cite{elman1990finding} in long sequence tasks through a series of optimizations such as complex diagonalization and linear recursion. The LRU’s architecture retains high-accuracy modeling of temporal stroke patterns while enabling parallel training.
	
	We conducted experiments on three subsets of the CASIA-OLHWDB1.1 dataset~\cite{liu2013online}: digits, English upper letters, and Chinese radicals. The proposed SW-PS+LRU framework consistently converged rapidly during training and achieved high test accuracy across all rotated subsets. The datasets and our code are publicly available.
	
\section{Related Work}
	The handwritten character recognition process can usually be divided into three main stages: data preprocessing, feature extraction and classification.
	Early dynamic time warping (DTW)~\cite{rakthanmanon2012searching} and principal component analysis (PCA)~\cite{abdi2010principal} can align and partially correct stroke trajectories. However, although DTW can compensate for temporal misalignment, it remains sensitive to large local distortions and does not provide rotation normalisation. While PCA can be effective for correcting rotated characters, it sometimes rotates the same character to different angles, making it harder to recognize. Path signature features (PSF)~\cite{graham2013sparse} characterize the global trajectory structure through iterative integration, which can improve the local detail representation by increasing the truncation order. However, excessively high orders will cause the feature dimension to increase sharply. To address this, Yang et al.~\cite{yang2016rotation} proposed a dyadic partition approach, in which the trajectory is recursively and evenly divided into $2^n$ segments at hierarchical level $n$, and path signature features are computed for each segment.
	In the classification stage, early approaches such as modified quadratic discriminant functions (MQDF)~\cite{liu2013online} and discriminative learning quadratic discriminant functions (DLQDF)~\cite{liu2004discriminative} achieved strong accuracy and stability, but were less effective under complex spatial rotations and often computationally expensive. With the success of deep convolutional neural networks (CNNs) in image recognition, handwritten character recognition gradually shifted toward end-to-end visual frameworks. Typical examples include CNN-Fujitsu~\cite{yin2013icdar} and multi-column deep neural networks (MCDNN)~\cite{cirecsan2015multi}, which significantly outperformed traditional classifiers in terms of accuracy. Models such as HCCR-GoogLeNet~\cite{zhong2015high} effectively integrate hand-crafted geometric features with convolutional networks. The model in~\cite{li2018rotation} further strengthens rotational robustness through a two-stage “rectify-then-recognize” procedure.

	In addition to static image features, online handwriting trajectories preserve the dynamic writing process, making sequence modeling a natural choice for recognition. Recurrent neural networks (RNNs)~\cite{zhang2017drawing}, including long short-term memory networks (LSTMs)~\cite{hochreiter1997long} and gated recurrent units (GRUs)~\cite{cho2014learning}, can capture sequential dependencies. However, their inherently serial computation limits parallelization and scalability, and their performance often degrades on long sequences~\cite{gu2021efficiently}. Transformers, the prevailing architecture for large language models, are highly effective but incur quadratic computational cost, making long-sequence modeling prohibitively expensive. On the other hand, State Space Models (SSMs)~\cite{gu2021efficiently} propagate information over time via parallel computation, enabling long-sequence modeling with linear complexity. State space model variants, such as LRU~\cite{orvieto2023resurrecting} and S5~\cite{smith2023simplified}, leverage simplified linear recurrence to effectively capture long-range dependencies. Neural Controlled Differential Equations (NCDE)~\cite{kidger2020neural} and Log Neural Controlled Differential Equations (Log-NCDE)~\cite{walker2024log} provide continuous-time representations for time series, but their reliance on numerical differential equation solvers results in high computational overhead, which limits their application.

\section{Methods}
	\subsection{Trajectory Preprocessing}
\begin{figure}
		\centering
		\includegraphics[width=0.85\textwidth]{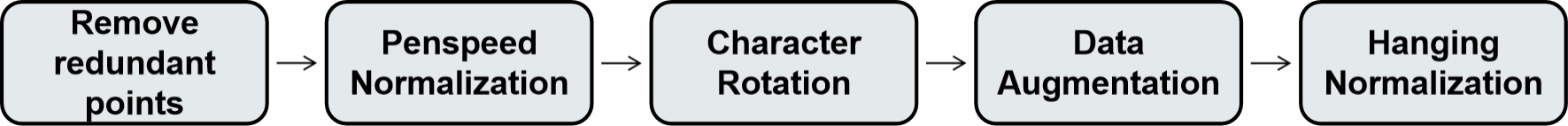}
			\vspace{-0.5em}
		\caption{Data preprocessing workflow.}
		\label{fig1}
	\end{figure}

	Based on~\cite{xiao2024integrating,pastor2005writing,simard2003best,du2012designing,yang2016rotation}, our data preprocessing pipeline consists of Removing Redundant Points, Penspeed Normalization, Character Rotation, Data Augmentation and Hanging Normalization. The workflow is shown in Fig.~\ref{fig1}.
	
	\vspace{-1.0em}
	\vspace{-1.5em}
				
	To preserve the temporal information of handwriting, we add an ink dimension~\cite{reizenstein2014signatures} $ink_i$, where $ink_i$ represents the writing order index. The input sequence is defined as \(S = [(x_i, y_i, ink_i, p_i)]_{i=1}^l\), where $(x_i, y_i)$ are coordinates and $ink_i$ is the writing order index. The pen state $p_i$ is a vector representing \emph{pen-down} $(0,1)$, \emph{continuous} $(0,0)$, or \emph{pen-up} $(1,0)$ states.
	
	To capture local dynamics, we compute first-order ($\Delta x, \Delta y$) and second-order ($\Delta^2 x, \Delta^2 y$) differences, representing velocity and acceleration. These differential features are linearly normalized to $(-1, 1)$ to eliminate scale bias.
	The final feature vector at each time step comprises $[x_i, y_i, ink_i, p_i, \Delta x_i, \Delta y_i, \Delta^2 x_i, \Delta^2 y_i]$. Trajectories are truncated or padded to a uniform length before the path signature extraction.
	
	\vspace{-1.0em}
	\subsubsection*{Character Rotation}
	Each handwritten sample is first rotated by a random angle $\alpha$ around the origin. The rotation transformation is defined as:
	\begin{equation*}
		\left[ x', y' \right] =
		[x, y]
		\left[ \begin{array}{cc}
			\cos \alpha &  \quad -\sin \alpha \\
			\sin \alpha &  \quad \cos \alpha
		\end{array} \right]
		\label{eq:rotation}
	\end{equation*}
	where $\alpha \in [0, 2\pi]$ represents the rotation angle.
	
	\vspace{-1.0em}
	\subsubsection*{Data Augmentation}
	We enhance the dataset's diversity through data augmentation, using affine transformations and elastic distortions~\cite{simard2003best}. First, random scaling and translation are applied:
	\[
	\left[ x', y' \right] =
	[x, y]
	\left[ \begin{array}{cc}
		1 + \delta_x & 0 \\
		0 & 1 + \delta_y
	\end{array} \right]
	+ \left[ \delta_{t_x}, \delta_{t_y}\right]
	\]
	where $(\delta_x, \delta_y)$ represent random scaling factors affecting the aspect ratio, and $(\delta_{t_x}, \delta_{t_y})$ are small translation jitters. Additionally, an elastic distortion is introduced:
	\[
	x'' = x' + \epsilon \sin ( 2\pi y' ), \quad y'' = y' + \epsilon \sin ( 2\pi x' )
	\]
	Here, $\epsilon$ controls the deformation strength. By enriching the spatial configuration of handwritten strokes, the augmented dataset better represents various natural writing variations encountered in practice.

	\vspace{-1.0em}
	\subsubsection{Hanging Normalization}
	To achieve a rotation-free representation and mitigate writing style differences, we employ the hanging normalization method~\cite{du2012designing,yang2016rotation}. Hanging normalization aligns two stable keypoints to a fixed vertical direction for rotational compensation before feature extraction. For the two keypoints, possible choices are the start point and center point (SC), the start point and end point (SE), and the average start point and average end point of all strokes (ASE). Following~\cite{yang2016rotation}, we adopt the SC mode and briefly describe its formula.

	Given an online handwritten character with $T$ time steps
	$P = \{ (x_t, y_t) \}_{t=1}^T$,
	we denote the starting point $S$ and the center point $C$ as:
	\[
		S = (x_1, y_1),\qquad
		C = (\bar{x}, \bar{y}) =
		\left( \frac{1}{T} \sum_{t=1}^{T} x_t,
		\frac{1}{T} \sum_{t=1}^{T} y_t \right)
	\]
	The compensation angle $\theta$ for rotating the line connecting $S$ and $C$ to the vertical direction is calculated as:
	\[
		\theta = \arctan \left( \frac{\bar{y} - y_1}{\bar{x} - x_1} \right) + \frac{\pi}{2}
	\]
	Then, each point is rotated by $-\theta$ around $S$ to obtain rotation-normalized coordinates.
	
\begin{figure}[htbp]
		\centering
		\includegraphics[width=\textwidth]{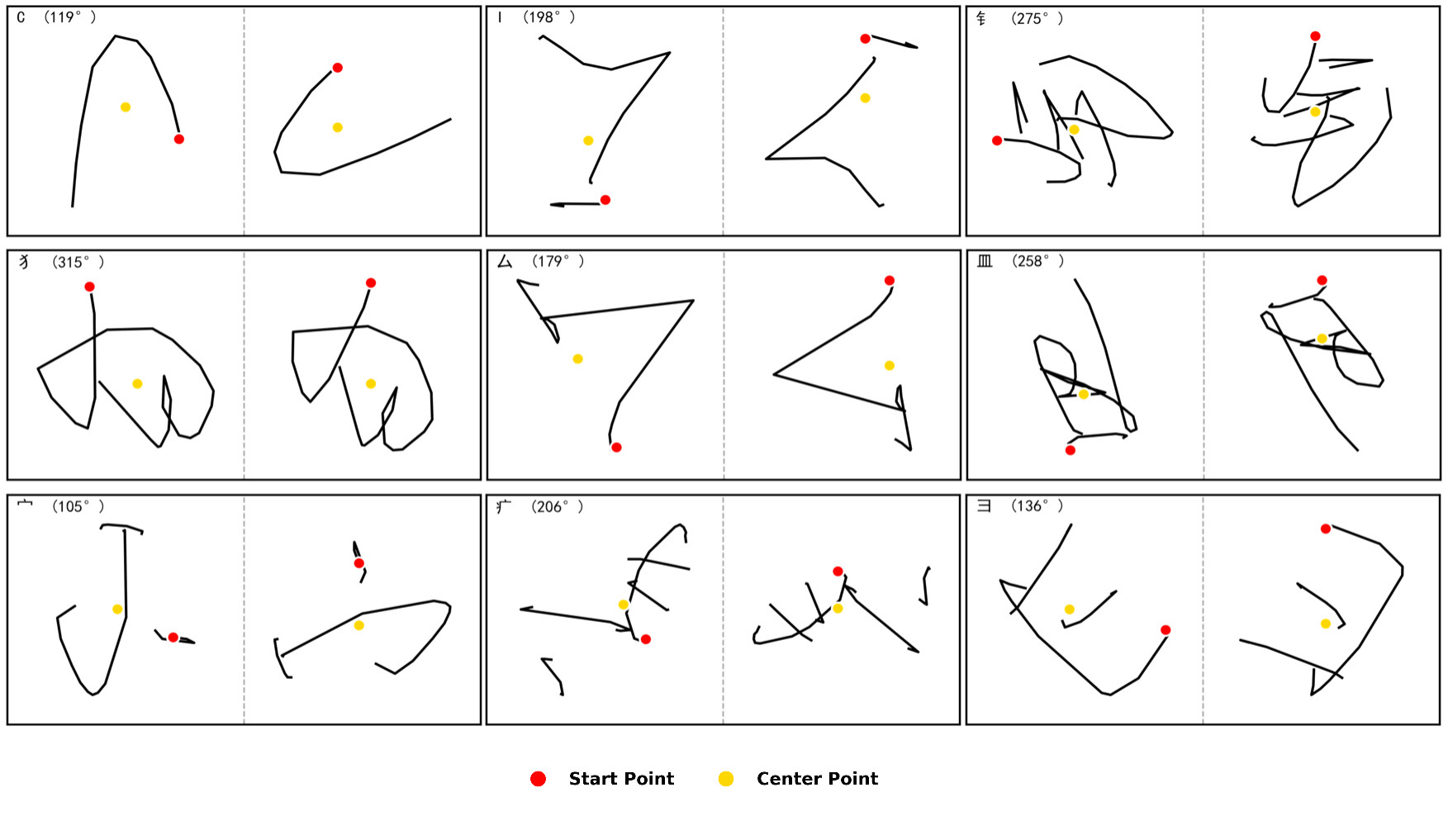}
				\vspace{-3.0em}
		\caption{Effect of Hanging Normalization in SC mode. The left side is "before", with the true label and rotation angle in the upper left corner. The right side is "after". The start red point is at the top and the center yellow point is at the bottom. Both are located on the same vertical line.}
		\label{fig2}
	\end{figure}

	Fig.~\ref{fig2} shows samples before and after hanging normalization.
	Hanging normalization largely removes rotation-induced directional variation and makes the stroke layout of the same character more consistent across rotation angles, thereby providing a more stable input for subsequent feature extraction and classification.
	
	\vspace{-1.0em}
	\vspace{-2.0em}

\subsection{Sliding Window Path Signature}
	\subsubsection{Path Signature}
\begin{figure}[htbp]
		\centering
		\includegraphics[width=0.7\textwidth]{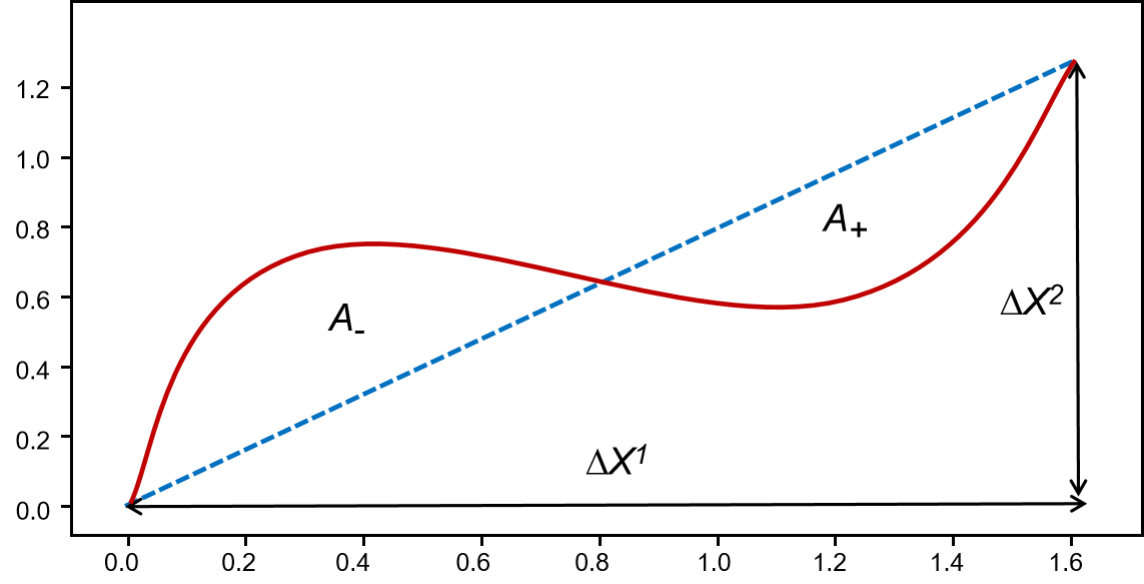}
		\vspace{-1.0em}
		\caption{Illustration of the step-2 truncated signature for a planar path (in red). The displacements $\Delta X^1$ and $\Delta X^2$ are the first level iterated integrals. The signed area $A_+-A_-$ is a linear combination of second iterated integrals based on Green's theorem.}
		\label{fig3}
	\end{figure}

	The path signature traces back to Chen~\cite{chen1958integration} in his work on geometric representation of paths, and was later systematically developed by Lyons in the Rough Path Theory~\cite{lyons1998differential}. It is obtained by computing iterated integrals on the coordinates of a path. This representation provides a top-down description of the geometry of the path: the first-order terms correspond to the total displacement along each coordinate axis; second-order terms are related to the signed area enclosed by the path together with the line segment joining the end point to the starting point (as shown in Fig.~\ref{fig3}).

	A discrete time series X can be linearly interpolated into a continuous trajectory:
	\[
	X: [0,T] \rightarrow \mathbb{R}^d, \quad X_t = \left( X_t^{1}, X_t^{2}, \dots, X_t^{d} \right).
	\]
	For a positive integer $k \geq 1$ and an index tuple $(i_1, \dots, i_k)$ with $i_j \in \{1,2,\dots,d\}$, the $k$-th iterated integral along $X$ is defined as:
	\[
	S^{i_1,\dots,i_k}_{0,T}(X) = \int_{0 < t_1 < \dots < t_k < T} dX_{t_1}^{i_1} \cdots dX_{t_k}^{i_k}.
	\]
	
	The step-$m$ truncated signature of a path $X$ consists of all its iterated integrals up to level $m$, forming a finite sequence. It is defined as:
	\[
	\mathrm{Sig}^m(X) = \left(S_{0,T}^{1}(X),\dots, \;S_{0,T}^{d}(X), \dots, S_{0,T}^{i_1,\dots,i_m}(X),\dots , S_{0,T}^{d,\dots,d}(X)\;  \right)
	\]
	with dimension $\sum_{k=1}^{m} d^k.$
	
	\vspace{-1.5em}
	\vspace{-2.5em}
		
	\subsubsection{Sliding Window Path Signature}
	Let the sequence $X$ be ${(s_1, s_2, \dots, s_L)}$. We segment the sequence into overlapping windows where $w$ is the window length and $t$ is the step size~\cite{graham2013sparse}:
	\[
	w_j = (s_{(j-1)t+1}, s_{(j-1)t+ 2}, \dots, s_{(j-1)t + w }), \quad j = 1, 2,\dots,K
	\]
	Each window $w_j$ is transformed using the $m$-th order signature mapping:
	\[
	SW_j = \mathrm{Sig}^m(w_j)
	\]
	The total number of windows is given by:
	\[
	K = \left\lfloor \frac{L - w}{t} \right\rfloor + 1
	\]
	where $\lfloor \cdot \rfloor$ denotes the integer floor operation.
	
	This windowed signature representation preserves sensitivity to local stroke deformations
	(e.g. elastic distortions) while maintaining the global series structure,
	making it well-suited for downstream sequence models.
	
\subsection{Linear Recurrent Unit (LRU)}
	In sequence modeling, traditional recurrent neural networks (RNNs) often encounter efficiency and stability issues during training, such as sequential processing constraints and vanishing or exploding gradients~\cite{bengio1994learning}.
	Linear Recurrent Neural Networks have emerged as a promising alternative, addressing the training efficiency issue by employing parallelizable linear recurrence. In particular, Linear Recurrent Unit (LRU) is introduced in~\cite{orvieto2023resurrecting}:
	\[
	h_t = A h_{t-1} + B x_t,\quad
	y_t = C h_t + D x_t,
	\]
	where $A, B, C, D$ are learnable parameters.
	
	With careful initialization and parameterization, LRU achieves  strong performance on the Long Range Arena benchmark~\cite{orvieto2023resurrecting}. To achieve efficient parallelizable training, LRU utilizes complex diagonal matrix $A$, allowing convolutional unrolling to compute the hidden representation $h_k = \sum^{k-1}_{j=0}A^jBx_{k-j} $ with $h_0 = Bx_0$. Therefore, using the eigenvalue decomposition of a complex diagonalizable matrix $A$, we can write:
	\[
	A = P \Lambda P^{-1}, \quad \Lambda = \mathrm{diag}(\lambda_1, \dots, \lambda_{d_h})
	\]
	
	Since real matrices that are complex-diagonalizable are dense~\cite{hartfiel1995dense}, any $n$-by-$n$ real matrix can be approximated by this decomposition. The diagonalization reduces matrix powers to element-wise operations $A^k = P \Lambda^k P^{-1}$, significantly boosting efficiency. The diagonal eigenvalues $\lambda_i$ are parameterized in the complex exponential form:
	\[
	\lambda_i = e^{-\nu_i + j \theta_i}, \quad \nu_i > 0
	\]
	Here, $\theta_i$ captures periodic patterns, while $\nu_i$ controls the exponential decay rate. The constraint $\nu_i > 0$ ensures $|\lambda_i| < 1$, preventing the hidden state from exploding over long sequences and guaranteeing long-term stability~\cite{orvieto2023resurrecting}.
	
	\vspace{-1.0em}
	\begin{figure}[H]
		\centering
		\includegraphics[width=\textwidth]{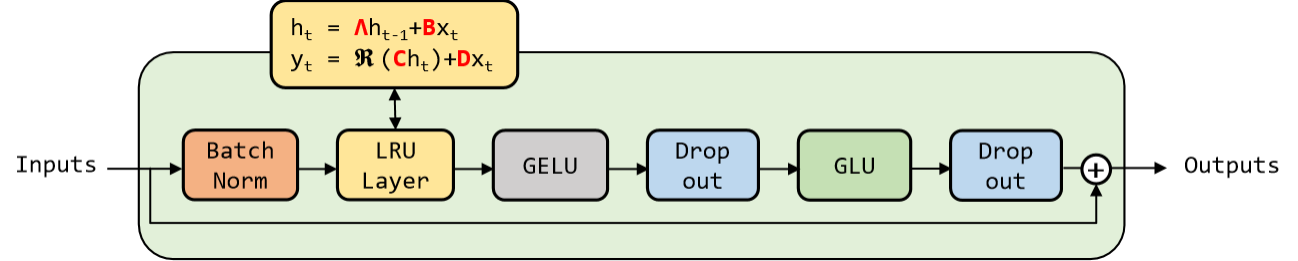}
		\vspace{-1.5em}
		\caption{The design of LRU Block. The Batch Norm mitigates covariate bias. The LRU Layer performs complex-valued linear recursion. Gaussian Error Linear Unit (GELU) and Gated Linear Unit (GLU) provide nonlinearities through smooth activation and learnable gating respectively. Dropout is applied between activations as a stochastic regularizer to mitigate overfitting and enhance generalization.}
		\label{fig4}
	\end{figure}
	\vspace{-2.5em}
	
\subsubsection{Overall Architecture of the Model} 
\begin{figure}[htbp!]
		\centering
		\includegraphics[width=1.0\textwidth]{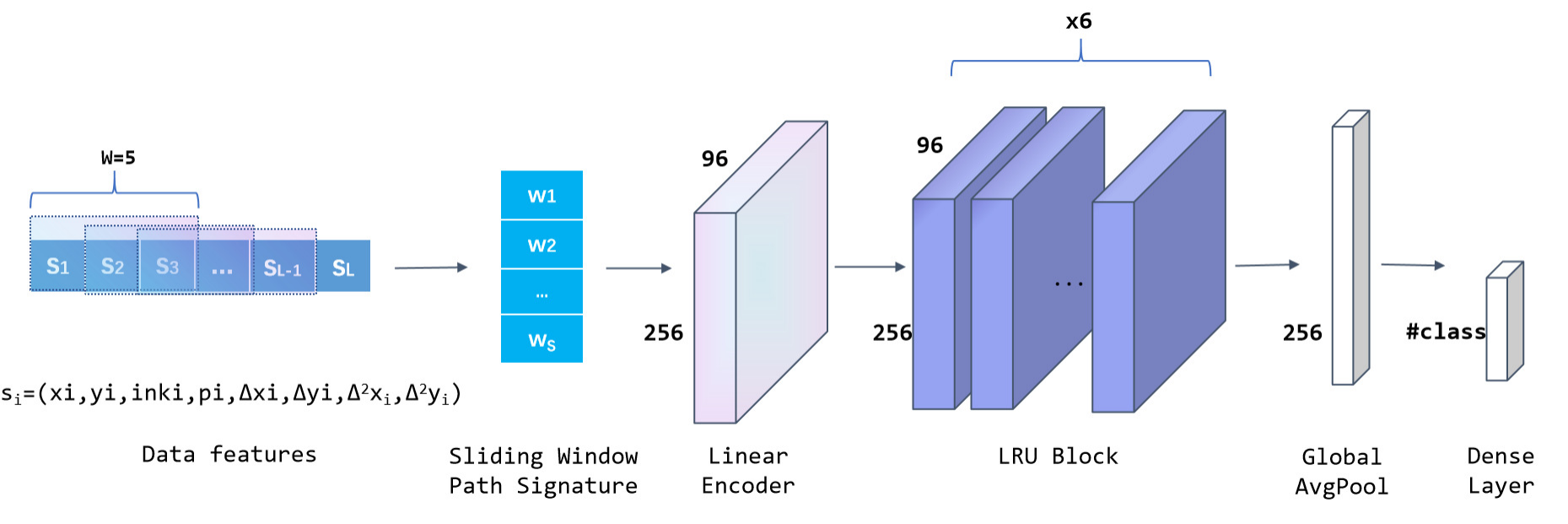}
			\vspace{-1.5em}
		\caption{Overall Architecture of the Model.}
		\label{fig5}
	\end{figure}

After data preprocessing and SW-PS feature construction, we perform sequence classification based on LRU. As shown in Fig.~\ref{fig5}, our model consists of three parts:

	\medskip
	\noindent\textit{Linear Encoder Layer}---Mapping the 90-dimensional SW-PS features of each window into a 256-dimensional hidden space.
	
	\medskip
	\noindent\textit{Stacked LRU Blocks}---The number of windows and the dimension of hidden layers remain constant in the stacked modules. As shown in Fig.~\ref{fig4}, each block consists of three parts:
	
	(1) Linear recurrence. The input is batch normalized and then fed to a complex-valued recursive LRU layer \(h_t = \Lambda h_{t-1} + Bx_t\);
	
	(2) Non-linear gating. The inherent linearity of LRU is compensated through a composite gating path: \(\mathrm{GELU} \to \mathrm{Dropout} \to \mathrm{GLU} \to \mathrm{Dropout}\);
	
	(3) Residual connection. The input is added to the output to facilitate gradient flow.
	
	\medskip
	\noindent\textit{Output Layer}---Features from all windows are reduced in dimension using global average pooling, and the class probabilities are generated by the final dense layer.
		
	\vspace{-1.5em}
	\vspace{-2em}

\section{Experiments}
\subsection{Experimental Setups}
	Experiments were conducted on three subsets of CASIA-OLHWDB1.1: Digits (10 classes), English Upper Letters (26 classes), and Chinese Radicals (52 classes). Each class contains about 300 samples, of which 240 are used for training and 60 for testing. During training, we apply random global rotations together with affine perturbations and elastic distortions for data augmentation. During testing, each sample is rotated at intervals of $12^\circ$, yielding 30 rotated variants per sample.
	
	We use the Adam optimizer with global gradient clipping. The initial learning rate is set to $10^{-3}$ and decayed by a factor of 0.5 every 4500 steps to a minimum of $10^{-6}$. Cross-entropy loss with L2 regularization is adopted, and the dropout rate is set to 0.3.

\subsection{Evaluation of SW-PS and Hanging Normalization}
	This section evaluates the impact of window size, truncation degree, and hanging normalization. Since the Chinese radical dataset is more challenging than the other two, we selected it for these experiments. Preliminary tests indicated that stride had negligible impact, so we fixed it to 1.
	
	Due to memory overflow with degree-3 path signatures, we limited the comparison to degree-1 and degree-2 signatures, as shown in Table~\ref{table1}. Excessively large windows fail to capture local details, whereas a window size of 1 prevents path signatures from capturing nonlinear interactions across channels. Therefore, we evaluated window sizes of 2, 5, 10, 15, and 20 using LRU as the downstream model.
	
	As shown in Table~\ref{table1}, hanging normalization consistently improves accuracy by 3\% to 4\% across hyperparameter combinations. Degree-2 signatures also generally outperform degree-1 signatures. The configuration with a window size of 5, truncation degree 2, and hanging normalization achieved the highest recognition accuracy of 93.37\%. Consequently, we adopt this configuration for subsequent experiments.
	
	\vspace{-1.0em}
	\begin{table}[ht]
		\centering
		\renewcommand{\arraystretch}{1.2}
		\setlength{\tabcolsep}{6pt}
		\caption{Test accuracy with different sliding window sizes $w$, truncation degrees $m$, and hanging normalization on the Chinese radical dataset.}
		\label{table1}
		\begin{tabular}{cccccccc}
			\toprule
			\multicolumn{3}{c}{\multirow{2}{*}{Test Accuracy (\%)}} & \multicolumn{5}{c}{Window Size $w$} \\
			\cmidrule(lr){4-8}
			\multicolumn{3}{c}{} & 2 & 5 & 10 & 15 & 20 \\
			\midrule
			
			\multirow{4}{*}{\makecell{Truncation \\ Degree $m$}}
			& 1
			& \multirow{2}{*}{\makecell{No Hanging \\ Normalization}}
			& 88.89 & 89.36 & 86.86 & 85.75 & 85.17 \\
			
			& 2
			&
			& 90.01 & \textbf{90.51} & 89.06 & 88.23 & 87.33 \\
			
			\cmidrule(lr){2-8}
			
			& 1
			& \multirow{2}{*}{\makecell{Hanging \\ Normalization}}
			& 92.85 & 93.21 & 93.08 & 92.78 & 92.24 \\
			
			& 2
			&
			& 92.92 & \textbf{93.37} & 93.26 & 93.24 & 92.88 \\
			\bottomrule
		\end{tabular}
	\end{table}
	\vspace{-1.0em}
		
\subsection{Evaluation of Different Models}
	This section compares several downstream models, including LRU~\cite{orvieto2023resurrecting}, S5~\cite{smith2023simplified}, RFOLHCR~\cite{yang2016rotation}, NCDE~\cite{kidger2020neural}, Log-NCDE~\cite{walker2024log}, Transformer~\cite{vaswani2017attention}, and ResNet-18~\cite{he2016deep}. RFOLHCR is our reproduction of the method proposed in~\cite{yang2016rotation}. LRU, S5, and RFOLHCR are evaluated using SW-PS features. NCDE and Log-NCDE are included as representative continuous-time models. Due to memory constraints, NCDE and Log-NCDE are trained on normalized raw trajectories rather than SW-PS features, which partly explains their inferior performance. Transformer and ResNet-18 are further introduced as sequence-based and image-based baselines, respectively. Transformer takes the processed trajectory as input, whereas ResNet-18 classifies fixed-size images generated from normalized trajectories after interpolation and padding. For a fair comparison, all samples are hanging-normalized before being fed into the corresponding classifiers. To reduce the effect of random initialization, each experiment is repeated over 10 random seeds, and the results are reported as mean test accuracy $\pm$ standard deviation.
	
	Based on Table~\ref{table2}, the LRU model consistently achieves the highest test accuracy on all three datasets, outperforming the results reported in~\cite{yang2016rotation}. S5 and ResNet-18 follow, while NCDE and Log-NCDE perform the worst. These results indicate that the path signature effectively captures discriminative features of online handwritten trajectories. Our reproduction of RFOLHCR does not reach the accuracy reported in the original paper~\cite{yang2016rotation}, likely due to differences in data preprocessing and experimental settings. Although Transformer and ResNet-18 achieve competitive performance, they still underperform compared with LRU and S5, further demonstrating the effectiveness of the proposed method.
		
	\vspace{-1.0em}	
	\begin{table}[htbp]
		\centering
		\renewcommand{\arraystretch}{1.2}
		\setlength{\tabcolsep}{8pt}
		\caption{Test accuracy of different downstream models (\%).}
		\label{table2}
		\begin{tabular}{lccc}
			\toprule
			Model & Digits & English Upper Letters & Chinese Radical \\
			\midrule
			LRU         & \textbf{99.17} $\pm$ 0.08 & \textbf{95.71} $\pm$ 0.19 & \textbf{93.37} $\pm$ 0.09 \\
			S5          & 99.15 $\pm$ 0.35 & 95.63 $\pm$ 0.21 & 93.19 $\pm$ 0.14 \\
			NCDE        & 97.50 $\pm$ 0.52 & 89.18 $\pm$ 0.55 & 84.69 $\pm$ 0.43 \\
			Log-NCDE    & 97.72 $\pm$ 0.21 & 90.59 $\pm$ 0.24 & 85.14 $\pm$ 0.38 \\
			RFOLHCR     & 98.07 $\pm$ 0.44 & 93.21 $\pm$ 0.22 & 88.47 $\pm$ 0.21 \\
			Transformer & 97.32 $\pm$ 0.35 & 94.83 $\pm$ 0.25 & 88.95 $\pm$ 0.52 \\
			ResNet-18   & 97.33 $\pm$ 0.21 & 94.94 $\pm$ 0.18 & 90.80 $\pm$ 0.16 \\
			\bottomrule
		\end{tabular}
	\end{table}
	\vspace{-1.0em}

	Table~\ref{table3} shows the results of ensemble learning on models trained with 10 random seeds. We evaluated both soft voting and hard voting.
	Soft voting averages predicted probabilities, whereas hard voting selects the class that receives the most predicted labels. The results confirm that ensemble learning improves accuracy, with soft voting yielding an approximately 1$\%$ increase in test accuracy on the challenging Chinese radical dataset.
	
	\vspace{-1.0em}
	\begin{table}[htbp]
		\centering
		\renewcommand{\arraystretch}{1.2}
		\setlength{\tabcolsep}{6pt}
		\caption{Soft vs. Hard voting ensemble test accuracy (\%).}
		\label{table3}
		\begin{tabular}{lcccccc}
			\toprule
			\multirow{2}{*}{Model}
			& \multicolumn{2}{c}{Digits}
			& \multicolumn{2}{c}{English Upper Letters}
			& \multicolumn{2}{c}{Chinese Radical} \\
			\cmidrule(lr){2-3} \cmidrule(lr){4-5} \cmidrule(lr){6-7}
			& Soft & Hard & Soft & Hard & Soft & Hard \\
			\midrule
			LRU         & \textbf{99.62} & 99.61 & \textbf{96.67} & 96.58 & \textbf{94.33} & 94.30 \\
			S5          & 99.39 & 99.37 & 96.21 & 95.99 & 93.89 & 93.84 \\
			NCDE        & 98.17 & 98.00 & 91.08 & 90.16 & 87.24 & 86.85 \\
			Log-NCDE    & 97.94 & 97.83 & 92.31 & 92.05 & 87.60 & 87.02 \\
			RFOLHCR     & 98.67 & 98.83 & 94.68 & 94.32 & 89.99 & 90.05 \\
			Transformer & 97.83 & 97.67 & 95.77 & 95.83 & 90.87 & 90.76 \\
			ResNet-18    & 98.00 & 97.83 & 95.13 & 95.06 & 92.04 & 91.59 \\
			\bottomrule
		\end{tabular}
	\end{table}
	\vspace{-1.0em}

	Table~\ref{table4} compares the training efficiency and model complexity of different downstream models in terms of time per 1000 training steps, average GPU memory usage, number of parameters, and FLOPs. RFOLHCR achieves the fastest training speed and the lowest FLOPs, indicating high computational efficiency. In contrast, NCDE and Log-NCDE require substantially longer training time despite having relatively few parameters, mainly due to their reliance on numerical differential equation solvers. LRU and S5 both deliver strong classification performance, while LRU is more efficient than S5 in parameter count and computational cost. Although Transformer and ResNet-18 use relatively little GPU memory, their FLOPs are considerably higher than those of the other models, especially ResNet-18. Overall, considering both test accuracy and computational efficiency, LRU offers the most favorable trade-off among the compared models.
	
	\vspace{-1.0em}
	\begin{table}[htbp]
		\centering
		\caption{Comparison of training efficiency and model complexity.}
		\label{table4}
		\renewcommand{\arraystretch}{1.2}
		\begin{tabular}{lcccc}
			\toprule
			Model & \makecell{Time per\\1000 Steps (s)} & \makecell{Avg. GPU Memory\\Usage (MB)} & \makecell{Parameters\\(M)} & \makecell{FLOPs\\(M)} \\
			\midrule
			LRU         & 48.96 & 4230.61 & 2.40  & 25.03   \\
			S5          & 45.62 & 4441.33 & 6.33  & 55.17   \\
			NCDE        & 81.25 & 829.21  & 0.21  & 19.82   \\
			Log-NCDE    & 75.21 & 2745.32 & 0.21  & 79.03   \\
			RFOLHCR     & 13.08 & 6240.54 & 1.12  & 2.25    \\
			Transformer & 52.16 & 592.75  & 3.17  & 631.73  \\
			ResNet-18   & 56.69 & 191.54  & 11.19 & 1113.08 \\
			\bottomrule
		\end{tabular}
	\end{table}
	\vspace{-1.5em}
		
\begin{figure}[htbp!]
		\centering
		\includegraphics[width=0.8\textwidth]{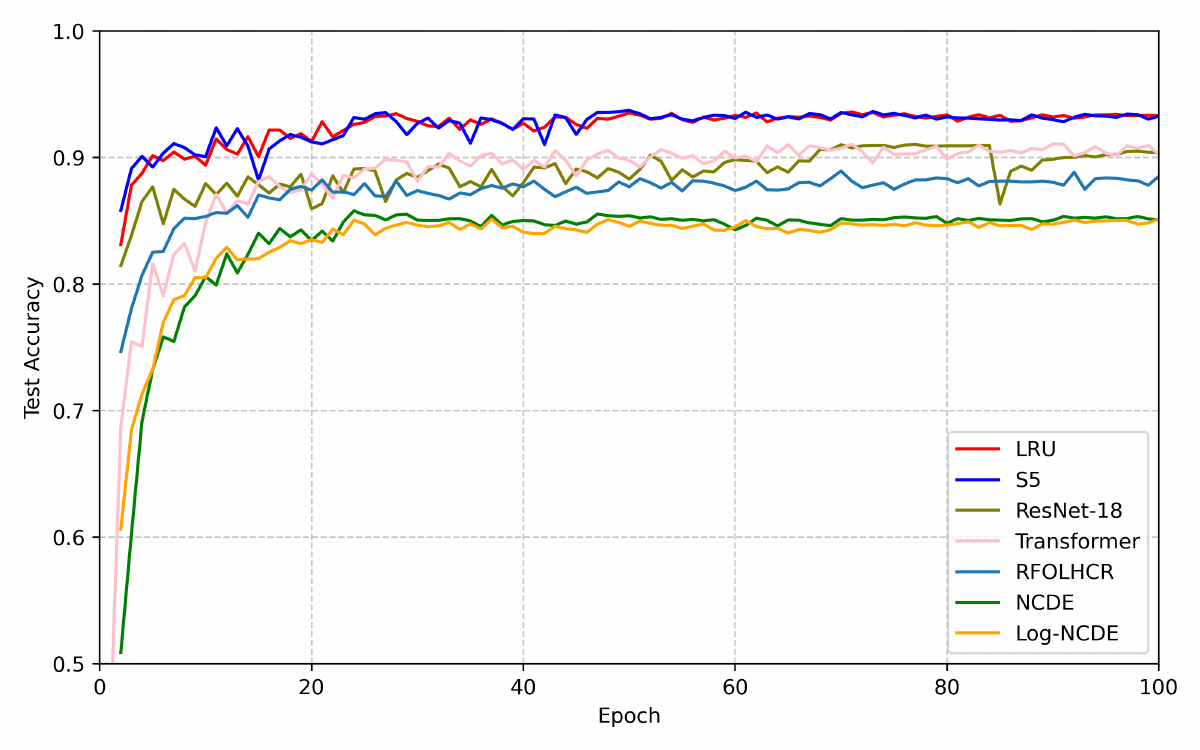}
		\vspace{-1.5em}
		\caption{Test accuracy on the Chinese Radical dataset (first 100 epochs).}
		\label{fig6}
	\end{figure}

	We further analyze the convergence behavior of different models. As shown in Fig.~\ref{fig6}, both LRU and S5 attain high test accuracy within the first 20 epochs and continue to improve thereafter. In contrast, the other baselines converge more slowly and generally plateau at lower accuracy levels. These results indicate that LRU and S5 are easier to optimise and exhibit higher training efficiency on the Chinese Radical dataset.
	
	\vspace{-1.0em}
	\vspace{-1.5em}
	
\begin{figure}[htbp!]
		\centering
		\includegraphics[width=0.9\textwidth]{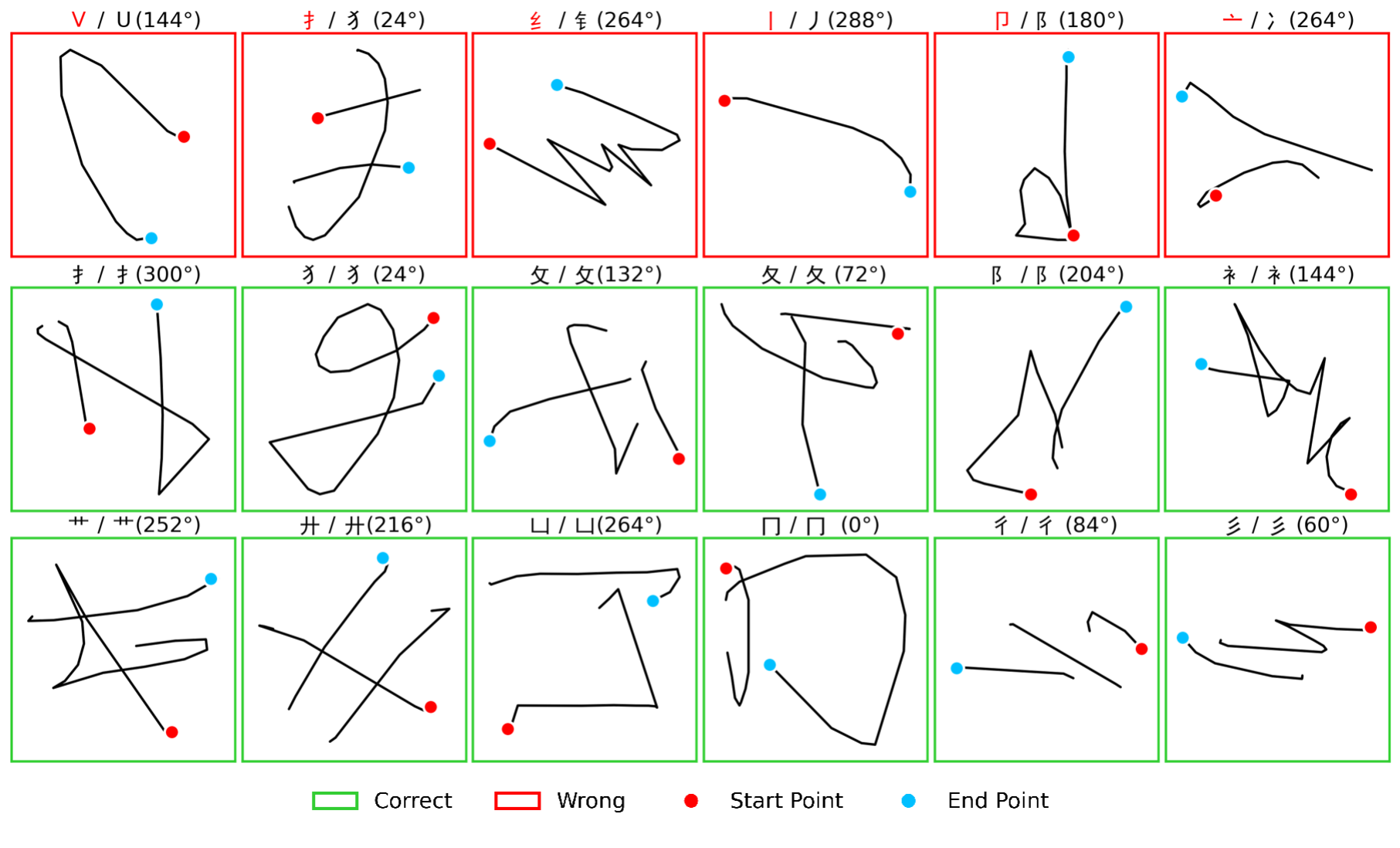}
		\vspace{-1.5em}
		\caption{Sample test results. The predicted class and true label are shown in the top left and top right, respectively, with the rotation angle given in brackets. The first row contains misclassified samples, whereas the second and third rows show correctly classified samples.}
		\label{fig7}
	\end{figure}

	\begin{CJK*}{UTF8}{gbsn}
		Table~\ref{table5} presents the results on the Chinese radical dataset under different rotation ranges. As the rotation range expands from $\pm 15^{\circ}$ to $\pm 180^{\circ}$, accuracy declines slightly. Large-degree rotations alter key structural features, causing confusion between similar radicals. For instance, in Fig.~\ref{fig7}, a “冫” rotated by $264^{\circ}$ geometrically resembles “亠”, leading to misclassification. In contrast, a “彡” rotated by $60^{\circ}$ retains its structural features and is correctly differentiated from “彳”. These samples illustrate that large-degree rotations introduce considerable ambiguity, whereas our proposed pipeline maintains high recognition accuracy despite these challenges.
	\end{CJK*}
	
	\vspace{-1.0em}
	\begin{table}[htbp!]
		\centering
		\renewcommand{\arraystretch}{1.2}
		\setlength{\tabcolsep}{5pt}
		\caption{Impact of rotation angle ranges on the Chinese radical dataset.}
		\label{table5}
		\begin{tabular}{llc}
			\toprule
			\textbf{Training Set} & \textbf{Testing Set} & \textbf{Test Accuracy (\%)} \\
			\midrule
			Non-rotated & Non-rotated & 94.63 \\
			Rotated within $\pm 15^{\circ}$ & Rotated within $\pm 15^{\circ}$ & 94.48 \\
			Rotated within $\pm 45^{\circ}$ & Rotated within $\pm 45^{\circ}$ & 93.54 \\
			Rotated within $\pm 180^{\circ}$ & Rotated within $\pm 180^{\circ}$ & 93.25 \\
			\bottomrule
		\end{tabular}
	\end{table}
	\vspace{-1.0em}
	
	\begin{CJK*}{UTF8}{gbsn}
	Finally, Fig.~\ref{fig7} shows some test results. The first row contains misclassified samples: a rotated “U” is mistaken for “V”, “犭” for “扌”, and “阝” for “卩”. These characters are structurally similar and hard to distinguish even without rotation. In the second and third rows, our method correctly classifies these ambiguous characters, demonstrating strong discriminative ability. Some samples are difficult to distinguish even for humans.
	\end{CJK*}
	
	\vspace{-1.5em}
	\vspace{-2.0em}
	
\section{Conclusions}
	This paper proposes an OLHCR framework that integrates SW-PS features extraction with LRU-based classification. By leveraging path signature feature and the sequence modeling capabilities of LRU, our model effectively balances local stroke variations with the global trajectory structure. As a result, our model achieves rapid convergence and high test accuracy while maintaining computational efficiency. Future work could explore the application of our system to rotation-free Chinese character recognition on larger datasets such as CASIA-OLHWDB1.1, as well as to handwritten characters in other languages.
	
	\vspace{-1.0em}
	\subsection*{Acknowledgements.}
	DY gratefully acknowledges the support of the National Natural Science Foundation of China (Young Scholar Grant 12201081), Chongqing University (Starting Grant 02080011044104, School of Mathematics and Statistics Basic Discipline Development Fund 0208005406001) and the Engineering and Physical Sciences Research Council (Programme Grant EP/S026347/1).
	
	\vspace{-1.0em}
	\subsection*{Data and Code Availability.}
	The data is publicly available at~\url{https://doi.org/10.6084/m9.figshare.30759515}. Our code is available at~\url{https://github.com/ling17154/Character_Recognition}.

%
%
 \bibliographystyle{splncs04}
 \bibliography{MeteorID141}

\end{document}